\documentclass[conference]{IEEEtran}
\IEEEoverridecommandlockouts
\usepackage{cite}
\usepackage{amsmath,amssymb,amsfonts}
\usepackage{graphicx}
\usepackage{textcomp}
\usepackage{xcolor}
\usepackage{graphicx}
\usepackage{amsthm}
\usepackage{algorithm} 
\usepackage{algpseudocode} 
\usepackage{booktabs}
\newtheorem{theorem}{Theorem}
\usepackage{afterpage}
\def\BibTeX{{\rm B\kern-.05em{\sc i\kern-.025em b}\kern-.08em
    T\kern-.1667em\lower.7ex\hbox{E}\kern-.125emX}}

\usepackage{fancyhdr}
\begin{document}

\title{CauSkelNet: Causal Representation Learning for Human Behaviour Analysis}

\author{\IEEEauthorblockN{Xingrui Gu}
\IEEEauthorblockA{
\textit{Helen Wills Neuroscience Institute} \\
\textit{University of California, Berkeley}\\
Berkeley, CA \\
xingrui\_gu@berkeley.edu}
\and
\IEEEauthorblockN{Chuyi Jiang}
\IEEEauthorblockA{
\textit{Department of Electrical Engineering} \\
\textit{Columbia University}\\
New York, NY \\
cj2792@columbia.edu}
\and
\IEEEauthorblockN{Erte Wang}
\IEEEauthorblockA{
\textit{Department of Computer Science} \\
\textit{University College London}\\
London, UK \\
ucabew0@ucl.ac.uk}
\and
\IEEEauthorblockN{Qiang Cui}
\IEEEauthorblockA{
\textit{The Future Laboratory} \\
\textit{Tsinghua University}\\
Beijing, China \\
cuiqiang@mail.tsinghua.edu.cn}
\and
\IEEEauthorblockN{Leimin Tian}
\IEEEauthorblockA{
\textit{School of Electrical and Computer Systems Engineering} \\
\textit{Monash University}\\
Melbourne, Australia \\
leimin.tian@monash.edu}
\and
\IEEEauthorblockN{Lianlong Wu}
\IEEEauthorblockA{
\textit{Department of Computer Science} \\
\textit{University of Cambridge}\\
Cambridge, UK \\
lianlong.wu@@cl.cam.ac.uk}
\and
\IEEEauthorblockN{Siyang Song}
\IEEEauthorblockA{
\textit{Department of Computer Science} \\
\textit{University of Exeter}\\
Exeter, UK \\
siyang.song0505@gmail.com}
\and
\IEEEauthorblockN{Chuang Yu}
\IEEEauthorblockA{
\textit{UCL Interaction Centre} \\
\textit{University College London}\\
London, UK \\
alexchauncy@gmail.com}
}
\maketitle
\thispagestyle{fancy}
\begin{abstract}

Traditional machine learning methods for movement recognition often struggle with limited model interpretability and a lack of insight into human movement dynamics. This study introduces a novel representation learning framework based on causal inference to address these challenges. Our two-stage approach combines the Peter-Clark (PC) algorithm and Kullback-Leibler (KL) divergence to identify and quantify causal relationships between human joints. By capturing joint interactions, the proposed causal Graph Convolutional Network (GCN) produces interpretable and robust representations. Experimental results on the EmoPain dataset demonstrate that the causal GCN outperforms traditional GCNs in accuracy, F1 score, and recall, particularly in detecting protective behaviors. This work contributes to advancing human motion analysis and lays a foundation for adaptive and intelligent healthcare solutions.
\end{abstract}


\section{Introduction}

Over the decades, traditional kinematic studies, using techniques like motion capture and biomechanical modeling, have advanced our understanding of human movement \cite{madeleine2008changes}\cite{james2010effects}. While valuable for analyzing body segment dynamics in areas like rehabilitation and performance evaluation, these methods often rely on observable metrics, such as joint angles and forces, limiting their ability to capture complex inter-joint dynamics. Recently, machine learning has transformed sports science by enabling advanced analysis of complex behavioral datasets \cite{olugbade2024emopain}\cite{romeo2023multi}. Techniques like deep learning and reinforcement learning have improved behavioral analysis and expanded exploration of sports techniques, addressing challenges such as modeling multidimensional joint interactions and uncovering their causal relationships.

Human behavior is a complex interplay of physiological and psychological dimensions shaped by emotions and cognition \cite{olugbade2019relationship}. Traditional frameworks offer foundational insights but often neglect the causal, bidirectional relationships between behavioral components. Research indicates that behavioral interactions follow Bayesian causal frameworks influenced by emotional and physiological states \cite{olugbade2019relationship}\cite{rachlin1985pain}. This underscores the need for advanced, interpretable algorithms to reveal the causal mechanisms driving behavioral dynamics, enabling a deeper understanding of physical and emotional interactions in real-world contexts.

While machine learning excels in movement recognition and prediction, many approaches prioritize classification accuracy over the causal relationships underlying behavioral dynamics \cite{khera2020role}. For instance, gait analysis often focuses on biomechanical classification, overlooking bidirectional joint interactions critical for movement optimization \cite{pfister2014comparative}. Understanding these causal links is key to improving model interpretability and enabling targeted interventions, paving the way for more personalized and adaptive applications in sports and healthcare.

This study introduces a novel approach combining KL divergence with causal inference to quantify causal relationships between human joints in behavioral contexts. While graph convolutional networks (GCNs) effectively represent inter-joint dynamics, their application in causal modeling remains underexplored. By integrating GCNs with causal inference, this research provides a robust framework for analyzing joint interactions, addressing existing limitations, and enabling advancements in personalized healthcare and sports performance optimization. The key contributions are summarized as follows:

\begin{itemize}
    \item \textbf{Causal Representation Framework:} A framework incorporating causal inference into human motion analysis to capture joint relationships and provide new insights into movement.
    
    \item \textbf{PC Algorithm and KL Divergence Integration:} A two-stage causal learning method combining the PC algorithm and KL divergence to uncover causal structures and quantify joint interactions, leading to more precise motion representations.
    
    \item \textbf{Interpretable and Robust Motion Recognition:} A motion recognition approach that generates interpretable causal graphs and demonstrates strong invariance to data scale changes, enhancing reliability and applicability in practical scenarios.
\end{itemize}

\section{Related Work}

\subsection{Traditional Kinesiology Research}
Traditional kinematic studies have advanced understanding of human movement using techniques like motion capture and biomechanical modeling to quantify dynamic changes in body segments. These methods have contributed significantly to fields such as clinical rehabilitation and sports performance evaluation \cite{james2010effects}\cite{madeleine2008changes}, revealing interactions between physical activities and physiological functions. However, they often focus on observable metrics, such as joint angles and force outputs, limiting their ability to model complex inter-joint dynamics. For instance, James et al. \cite{james2010effects} showed that fatigue alters biomechanical and neuromuscular responses during landing, while Cop et al. \cite{cop2021unifying} highlighted the difficulty of representing causal interactions between neural and biomechanical levels. While traditional methods provide a strong foundation, they struggle to capture the multidimensional factors influencing motor performance, underscoring the need for advanced representation learning techniques to model these complex dynamics more effectively.

\subsection{Multi-modal Data and Representation Learning}


Advancements in deep learning and multi-modal data modeling have greatly improved human behavior analysis by leveraging diverse data sources such as motion capture, biomechanical analysis, and physiological monitoring. Techniques like Graph Convolution Networks (GCNs) and Parallel Spatiotemporal Encoding Motion Representation (P-STEMR) have significantly advanced the modeling of movement dynamics \cite{olugbade2024emopain, cen2022exploring, yan2018spatial}. GCNs excel in capturing inter-joint interactions, while frameworks like HAR-PBD enhance behavior recognition by continuously monitoring joint dynamics \cite{cen2022exploring}. Despite these successes, most research emphasizes unidirectional influences, neglecting the complex causal relationships between behavior, emotion, and physiology \cite{pearl2009causal, scholkopf2022causality}. Integrating causal inference into these models could address this gap, improving the understanding of movement mechanisms. Hybrid approaches, such as combining GCNs with temporal models like LSTMs, have shown promise in capturing dynamic, multimodal relationships \cite{yan2018spatial, kipf2016semi}. These advancements underscore the need for models that effectively represent behavioral complexity while enabling personalized intervention strategies.


\section{Methodology}

To address the research gap in causal relationships among joints during specific behaviors, we employ causal graphical models \cite{nergui2014understanding,van2015cognitive,raita2021big,pearl2010causal} to explore causal structures within the human anatomical framework. We use Directed Acyclic Graphs (DAGs) with nodes representing joints and edges denoting causal interactions \cite{nergui2014understanding}. The graph's  topology is based on human physiology, with nodes at critical joints and edges mapping physical links. This approach captures the inherent connectivity among body segments and provides a foundation for detailed causal investigations into human pain responses and behavior.

Specifically, we employ the PC algorithm and Kullback-Leibler (KL) divergence to develop and validate causal graphical models, analyzing interactions among human joints\cite{kullback1951kullback,spirtes2001causation}. The PC algorithm identifies causal structures through conditional independence tests as:
\begin{equation}
   X \perp\!\!\!\perp Y \mid S
\label{PC eq} 
\end{equation}
where $X$ and $Y$ represent the state variables (e.g., position, velocity, or acceleration) of two joints in our model, and $S$ is a conditioning set of other joint states that might influence the relationship between $X$ and $Y$. This conditional independence test checks if $X$ and $Y$ are independent given $S$. If $X \perp\!\!\!\perp Y \mid S$ holds, it indicates no direct causal link between $X$ and $Y$ after accounting for the influence of $S$. This test is key to constructing the initial undirected graph in our causal model. For example, when analyzing the relationship between the knee joint ($X$) and ankle joint ($Y$), conditioning on the hip joint state ($S$) helps determine if their relationship is direct or mediated by the hip joint.

This algorithm is enhanced to establish a preliminary undirected graph, capturing latent connections among variables\cite{tsamardinos2006max, cieza2006identification, ramsey2012adjacency}. To determine causality directions, we employ Kullback-Leibler (KL) divergence, which quantifies the difference between probability distributions P and Q based on following Equation\ref{KL eq}:
\begin{equation}
    D_{\text{KL}}(P \parallel Q) = \sum_{x \in \mathcal{X}} P(x) \log\left(\frac{P(x)}{Q(x)}\right)
\label{KL eq1}
\end{equation}
  where $D_{\text{KL}}(P \parallel Q)$ measures the divergence between two distributions, $P(x)$ and $Q(x)$, representing the probability of a joint's state under conditions $P$ and $Q$, respectively. A larger $D_{\text{KL}}$ indicates a greater difference between the distributions, suggesting a stronger potential causal relationship. In our model, this is used to assess the asymmetric information flow between joints. For example, if $D_{\text{KL}}(P_i \parallel P_j)$ is significantly larger than $D_{\text{KL}}(P_j \parallel P_i)$, it suggests joint $i$'s movement pattern provides more information about joint $j$'s movement, implying a potential causal direction from $i$ to $j$.

To assess the influence strength between nodes and infer causal directions, for any two nodes $(i)$ and $(j)$, we calculate the KL divergence between $P(j \mid i)$ and $P(j)$, and vice versa. This reveals asymmetry in information flow, indicating potential causal directions. A larger KL divergence suggests reduced prediction uncertainty of node $(j)$ given node $(i)$, implying a likely causal direction. This method identifies direct connections and quantifies causal strengths, providing a robust foundation for establishing causal directions in DAGs. It extends to dynamically changing biomedical data, elucidating intricate interactions between movement behaviors and pain responses.

However, KL divergence has limitations: Firstly the sensitivity to sample size potentially affects causal direction determination. Next, as a non-symmetric metric, it illustrates improved prediction accuracy rather than direct causal strength. To address these issues, we employ statistical corrections and comparative analysis with additional techniques, namely information criteria and cross-validation, to refine causal direction assessments. These measures ensure more reliable causal inferences across different models and datasets.

\subsection{Formulation}






Let $\mathcal{G} = (V, E)$ be a graph representing the human body structure, where:
\begin{itemize}
    \item $V = \{v_1, \ldots, v_n\}$ denotes the set of nodes representing human joints, with $n$ being the total number of joints;
    \item $E \subseteq V \times V$ denotes the set of edges representing physical or causal connections between joints;
    \item $v_i \in V$ represents the $i$-th joint in our skeletal model.
\end{itemize}

Our objective is to derive a representation $\phi: V \rightarrow \mathbb{R}^d$ that encapsulates the causal relationships between joints, where $d$ is the dimension of our representation space. We employ the PC algorithm to elucidate the causal structure. For variables $X, Y \in V$ and a set of conditioning variables $S \subset V \setminus \{X, Y\}$, we examine the conditional independence with Eq.\ref{PC eq}. The PC algorithm yields a completed partially directed acyclic graph (CPDAG) $\mathcal{G}^* = (V, E^*)$, where $E^*$ represents the set of directed edges identifying causal relationships. To quantify the strength of causal relationships and derive joint representations, we utilize KL divergence. For any edge $(i, j) \in E^*$, we define:
\begin{equation}
    D_{KL}(i \rightarrow j) = \sum_{x_j \in \mathcal{X}_j} P(x_j \mid x_i) \log\left(\frac{P(x_j \mid x_i)}{P(x_j)}\right)
\label{KL eq}
\end{equation}
where:
\begin{itemize}
    \item $x_i, x_j$ represent the state variables of joints $i$ and $j$ respectively;
    \item $\mathcal{X}_j$ is the set of all possible states for joint $j$;
    \item $P(x_j \mid x_i)$ is the conditional probability of joint $j$ being in state $x_j$ given joint $i$ is in state $x_i$;
    \item $P(x_j)$ is the marginal probability of joint $j$ being in state $x_j$.
\end{itemize}
Subsequently, we formulate a representation function $\phi$ as:
\begin{equation}
    \phi(v_i) = \left[\frac{D_{KL}(i \rightarrow j)}{\sum_{k \in \text{Ne}(i)} D_{KL}(i \rightarrow k)}\right]_{j \in \text{Ne}(i)}
\label{rep eq}
\end{equation}
where:
\begin{itemize}
    \item $\text{Ne}(i)$ denotes the set of neighboring joints directly connected to joint $i$ in $\mathcal{G}^*$;
    \item The representation $\phi(v_i)$ is a vector whose components are normalized KL divergence values;
    \item Each component represents the relative strength of causal influence from joint $i$ to its neighboring joints.
\end{itemize}

\subsection{Theoretical Guarantees}
\begin{theorem}
    The representation $\phi$ preserves the causal structure of $\mathcal{G}^*$. (See Proof in Appendix \ref{proof1})
\end{theorem}

\begin{theorem}
    The representation $\phi$ is invariant to the scale of the data. (See Proof in Appendix \ref{proof2})
\end{theorem}

\begin{algorithm}
\caption{Causal Analysis Using PC Algorithm and KL Divergence}
\label{alg:causal_analysis}
\begin{algorithmic}[1] 
    \State \textbf{Initialization:} Construct graph skeleton $G=(V, E)$ from data $\mathcal{D}$ using PC algorithm.
    \State \textbf{Refinement:} Refine $G$ by adjusting edges $\epsilon \in E$ based on additional conditional independence tests.
    \State \textbf{DAG Formation:} Formulate DAG $\mathcal{G}=(V, \vec{E})$ by directing $E$ using dependencies derived from $\mathcal{D}$.
    \State \textbf{KL Divergence Calculation:}
    \State \hspace{\algorithmicindent} Estimate conditional probabilities $P(j|i)$ for all $i, j \in V$ using Maximum Likelihood Estimation.
    \State \hspace{\algorithmicindent} Estimate marginal probabilities $P(j)$ for all $j \in V$.
    \State \hspace{\algorithmicindent} For each pair $(i, j) \in \vec{E}$, compute:
    \State \hspace{\algorithmicindent}\hspace{\algorithmicindent} 
        $D_{\text{KL}}(i \rightarrow j) = \sum_{x_j \in \mathcal{X}_j} P(x_j | x_i) \log\left(\frac{P(x_j | x_i)}{P(x_j)}\right)$
    \State \hspace{\algorithmicindent}\hspace{\algorithmicindent} 
        $D_{\text{KL}}(j \rightarrow i) = \sum_{x_i \in \mathcal{X}_i} P(x_i | x_j) \log\left(\frac{P(x_i | x_j)}{P(x_i)}\right)$
    \State \textbf{Causal Direction Assessment:} Determine the direction of causality based on the asymmetry of the KL divergence values. A higher $D_{\text{KL}}$ from $i$ to $j$ compared to from $j$ to $i$ suggests that $i$ causally influences $j$.
    \State \textbf{Edge Weight Assignment:} Assign weights to the edges in the DAG based on the computed KL divergence values, reflecting the strength of the causal influence.
    \State \textbf{Model Validation:} Use statistical tests and cross-validation techniques to validate the causal model, ensuring robustness and reliability of the inferred causal relationships.
\end{algorithmic}
\end{algorithm}

\section{Experiment}

\subsection{Dataset}
\begin{figure}[htbp]
\centering
\includegraphics[width=8cm]{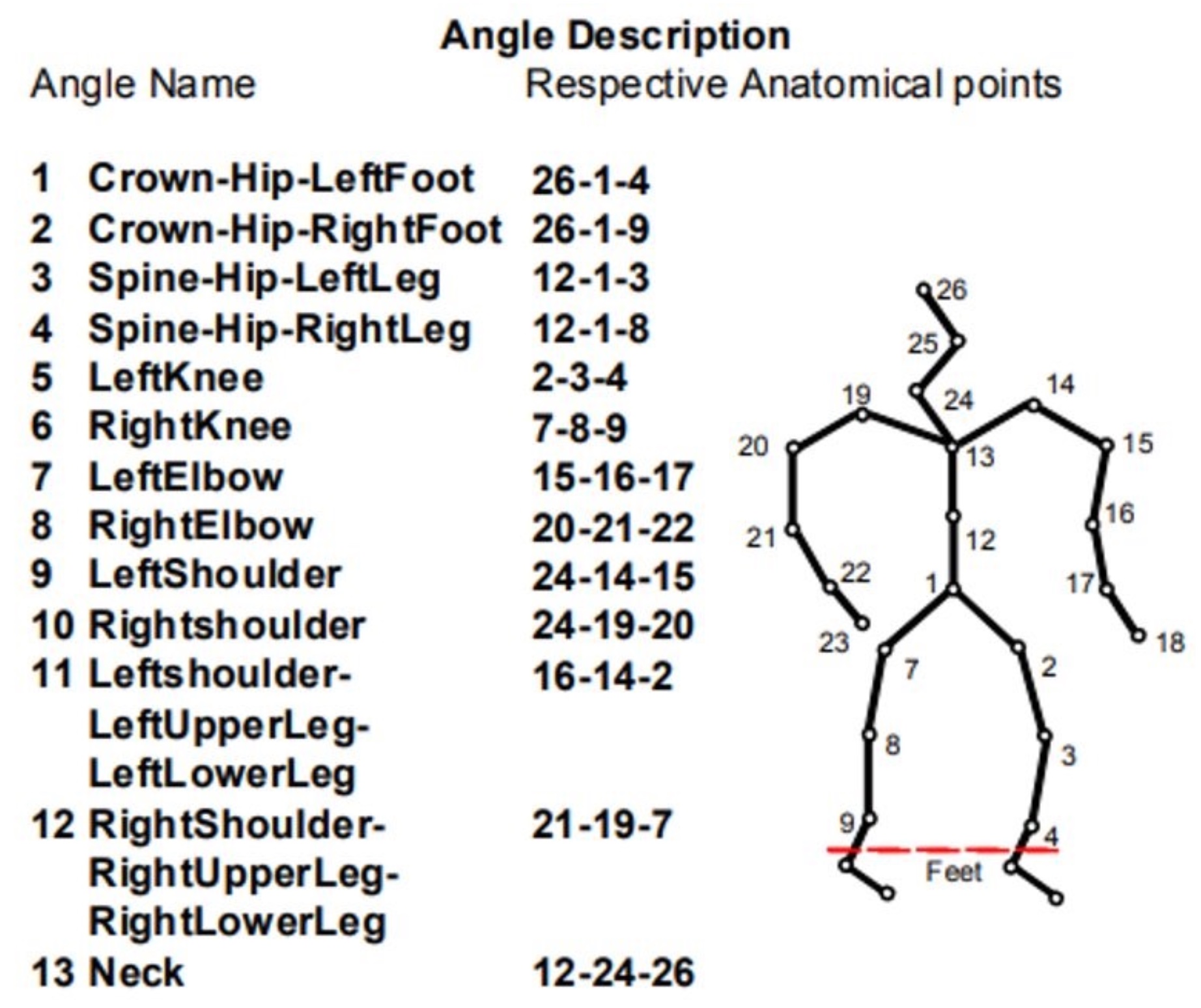}
\caption{Joint number and relative information in EmoPain Dataset}
\label{fig:emopain}
\end{figure}

The EmoPain dataset \cite{aung2015automatic} employed in this study is designed to analyze movement behaviors and pain recognition in chronic pain (CP) patients relative to a healthy control group~\cite{wang2021leveraging}. This dataset records pain levels and protective behaviors from 18 IMUs of 12 healthy and 18 CP participants within a controlled movement challenge, providing insights into how pain influences movement. As depicted in Figure \ref{fig:emopain}, the dataset constructs a fundamental human causal graph using the X, Y, and Z coordinates of 22 body joints, capturing body postures during movements and facilitating in-depth analyses of human motion patterns. EmoPain, as a publicly available dataset, provides detailed demographic information about the participants in its referenced articles\cite{aung2015automatic}. Specific details can also be found in Table \ref{tab:clbp_participants} in the Appendix.

\subsection{Experiment Design}

In this study, depicted in Figure \ref{fig:emopain}, we used the EmoPain dataset to create an undirected causal graph model of human structure, examining causal connections between joints in protective and non-protective behaviors and their impact on pain analysis. We processed the three-dimensional joint coordinate data (X, Y, Z) to capture dynamic interactions, then applied the PC algorithm to learn a DAG of potential causal relationships. A DAG was constructed based on the PC algorithm output, and the strength of causal relationships between joints was quantified using KL divergence measures. For each joint pair (e.g., Joint 1 and Joint 2), we calculated the conditional probability distribution and used KL divergence to measure information transfer efficiency, reflecting the reduction in uncertainty of one joint's state given another's. We compared KL divergence across dimensions to identify the dominant causal direction, selecting the most frequent direction and using its average KL divergence as a weight for information transfer efficiency. This analysis delved into joint interactions and their role in pain perception and protective behaviors, offering new insights for behavior-based pain analysis.

\subsection{Experiment Analysis}
\begin{figure}[ht]
\centering
\includegraphics[width=0.5\textwidth]{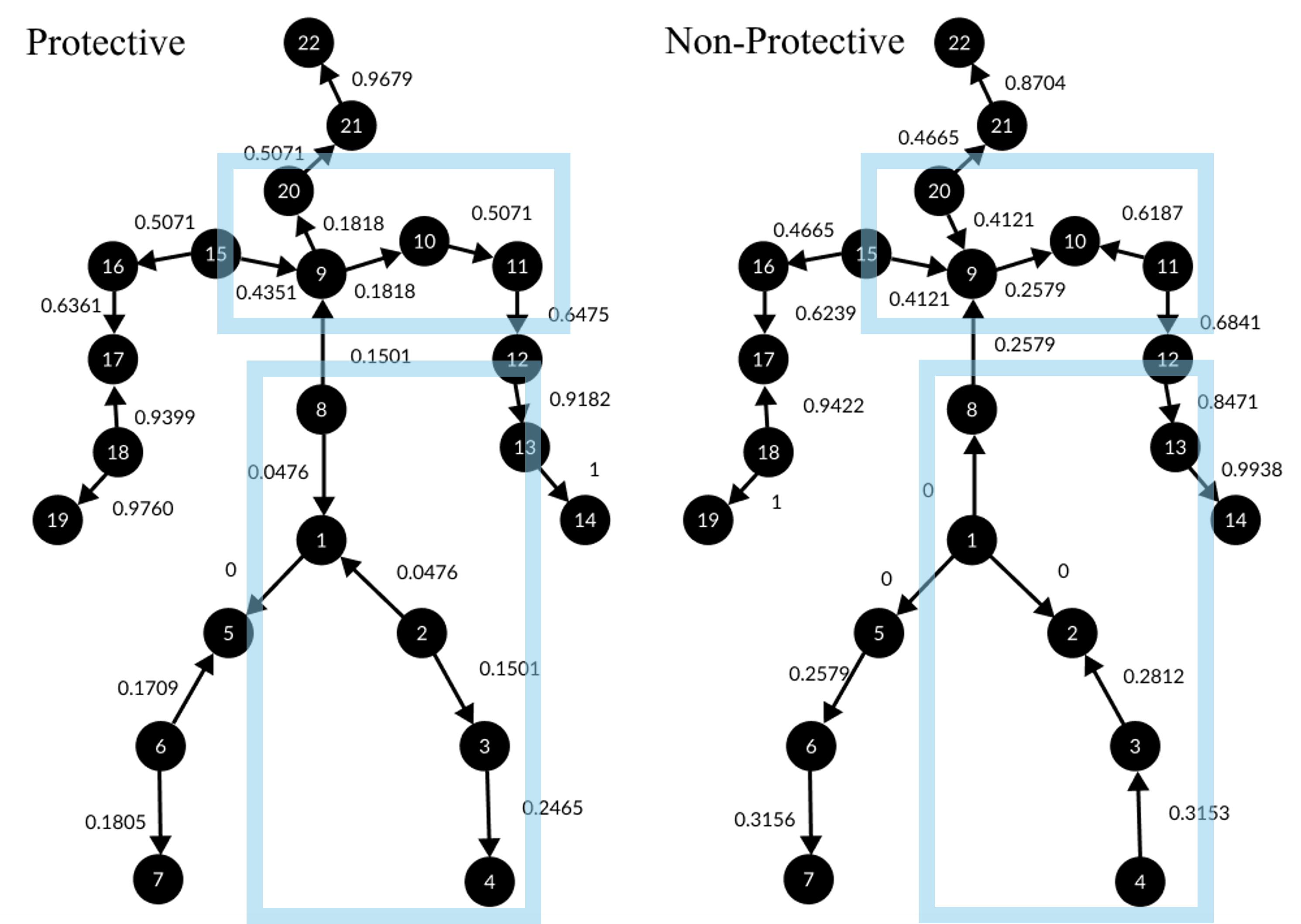}
\caption{The two different types of casual graphs for protective and non-protective behaviors. Some differences between the models could be found: e.g. edge between joint 9 and 20  and the edge weight} 
\label{fig:30000} 
\end{figure}

\begin{figure*}[ht]
  \centering
  \includegraphics[width=\textwidth]{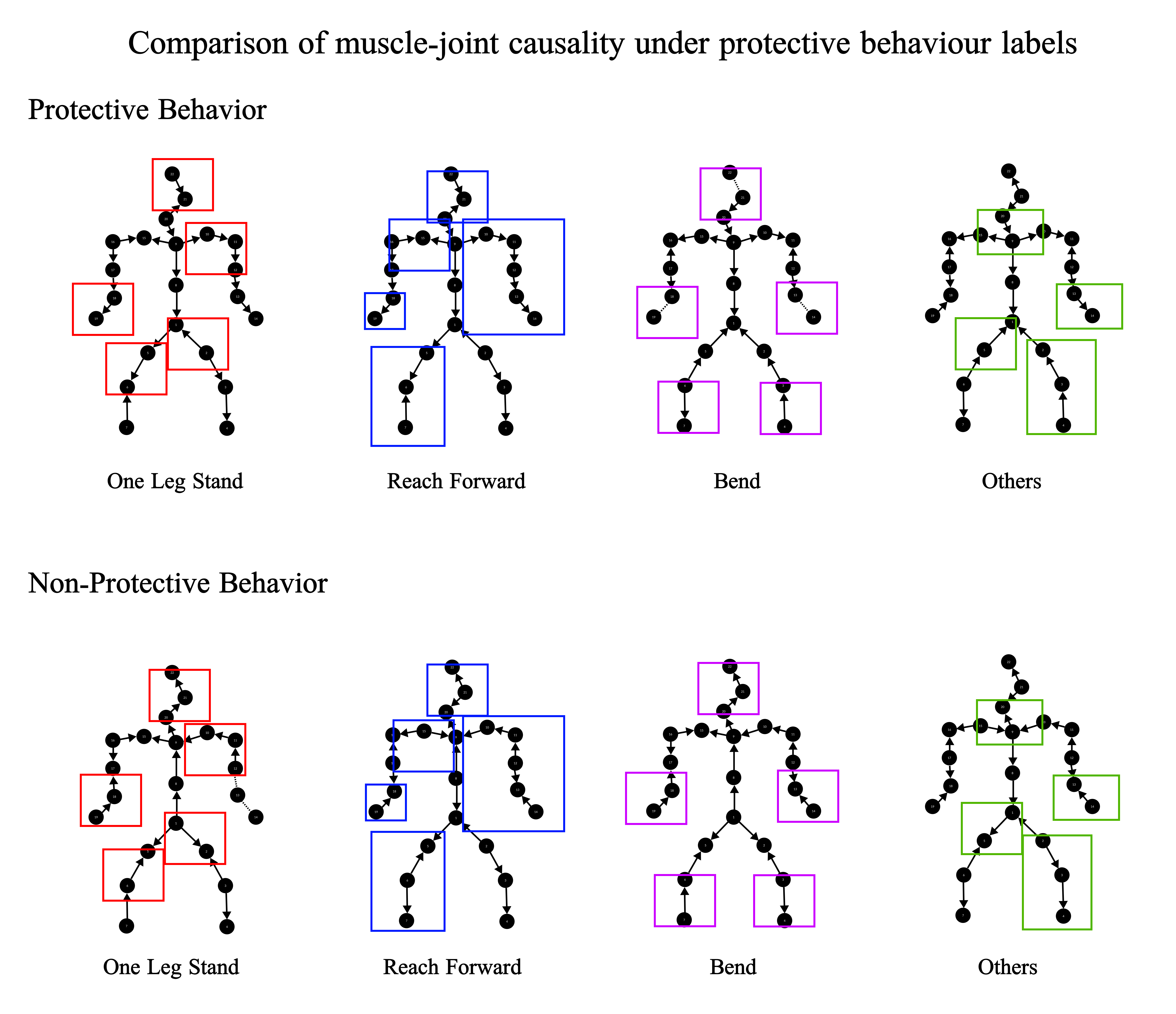}
  \caption{Causal graphs for four representative exercise types in both protective and non-protective behaviors are displayed, from left to right: One Leg Stand, Reach Forward, Bend, and Others. Each colored box highlights the differences between protective and non-protective behaviors based on graph representations, such as the direction and strength of the arrows. Detailed visualizations of all causal graphs for the nine movement types can be found in Appendix C.}
  \label{fig:exercisetype}
\end{figure*}
Causal graphs for protective and non-protective behaviors were plotted using 30,000 data samples, with Kullback-Leibler divergence normalized for comparison. Dashed lines indicate imprecise causal connections. As shown in Figure \ref{fig:30000}, significant differences between protective and non-protective behaviors were observed, such as reversed causal relationships and varied weights between certain nodes (e.g., nodes 9 and 20). These differences suggest that the human body adopts various physical adjustment strategies in response to pain. For instance, protective behaviors may restrict joint movements, while non-protective behaviors allow wider motion ranges. Similar variations were observed between other node pairs under different pain perceptions and emotional states. These findings provide insights into how pain influence human motor behavior, offering a bio-mechanical basis for personalized medical and adaptive machine learning models. In practical applications, this allows for more precise therapeutic recommendations and movement guidance based on individual behavior types.

To explore variations in joint interactions between protective and non-protective behaviors under different movement types, we plotted causal graph models labeled by exercise type, as shown in Figure \ref{fig:exercisetype}. By comparing causal graphs across different behavioral contexts, we observed consistency in the causal relationships among joints under similar pain perceptions and protective mechanisms. However, significant differences emerged when pain triggered a shift from non-protective to protective behaviors, demonstrating how pain perception and protective behaviors jointly influence human motion. These findings validate our hypothesis: under pain and emotional distress stimuli, causal connections among joints undergo notable changes as protective behaviors are prompted. In protective behaviors, to alleviate pain or prevent further injury, the activity of certain joints can be consciously restricted, reflected by changes in the direction and strength of causal relationships. Conversely, in non-protective behaviors, joints may exhibit greater freedom of movement, displaying different neural responses and behavioral adaptation to pain, with weaker causal constraints.

\begin{figure*}[htbp]
  \centering
  \includegraphics[width=\textwidth]{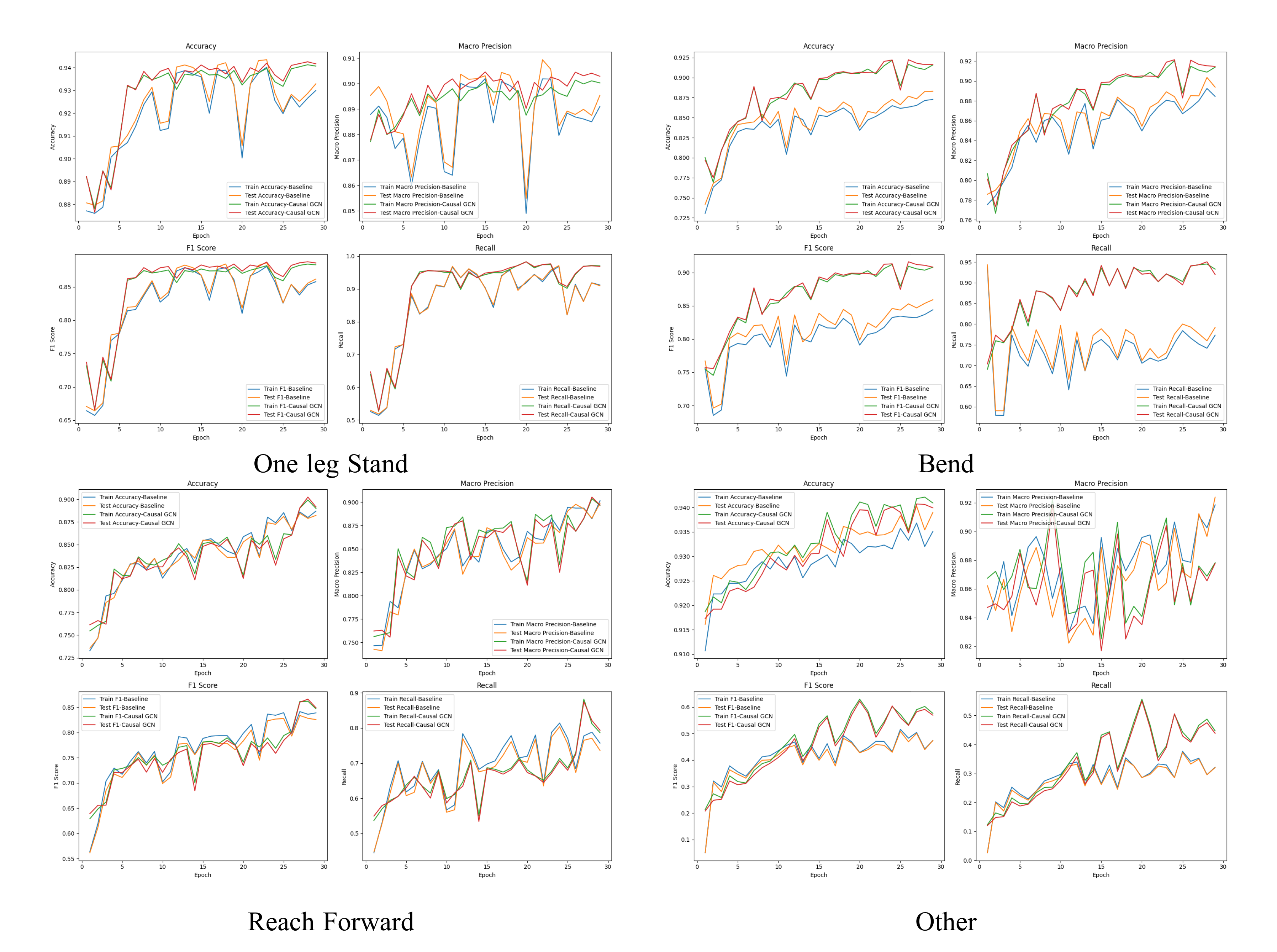}
  \caption{Comparative Trend Analysis of Performance Metrics Across Four Exercise Types for Baseline and Causal GCN Models.}
  \label{fig:metrics_trend_comparison}
\end{figure*}

\afterpage{%
\begin{table*}[htbp]
\centering
\caption{Performance Metrics for Baseline vs. Causal GCN in test}
\label{tab:metrics_comparison}
\small 
\setlength\tabcolsep{10pt} 
\begin{tabular}{@{}lcccc|cccc@{}}
\toprule
\textbf{Exercise Type} & \multicolumn{4}{c|}{\textbf{Baseline}} & \multicolumn{4}{c}{\textbf{Causal GCN}} \\ \midrule
 & \textbf{Acc.} & \textbf{Prec.} & \textbf{F1} & \textbf{Rec.} & \textbf{Acc.} & \textbf{Prec.} & \textbf{F1} & \textbf{Rec.} \\ 
\textbf{One Leg Stand} & 0.933 & 0.895 & 0.862 & 0.910 & \textbf{0.942} & 0.903 & 0.886 & \textbf{0.969} \\
\textbf{Bend} & 0.883 & 0.894 & 0.859 & 0.792 & 0.916 & 0.915 & \textbf{0.908} & 0.920 \\
\textbf{Reach Forward} & 0.882 & 0.899 & 0.826 & 0.736 & 0.892 & 0.897 & 0.849 & 0.794 \\
\textbf{Other} & 0.939 & \textbf{0.924} & 0.474 & 0.320 & 0.940 & 0.878 & 0.569 & 0.439 \\ \bottomrule
\end{tabular}
\end{table*}
}

\section{Evaluation}
\subsection{Evaluation Design}

In this study, we built upon our previous research involving causal directed acyclic graphs (DAGs). Our preliminary results showed that the causal graph model effectively captured changes in joint connections under different behavioral states. To advance this, we designed a comprehensive evaluation approach:
\begin{itemize}
    \item \textbf{Baseline Model:} We used an undirected graph based on human anatomy as our baseline, inspired by the success of Graph Convolutional Networks (GCN) in similar datasets \cite{cen2022exploring}. (See GCN model detail in Fig \ref{fig:gcnarticture})
    \item \textbf{Comparative Analysis:} We compared causal graph models of protective behaviors across various movement types with the baseline model, maintaining interpretability by avoiding edge weight additions.
    \item \textbf{Generalizability Test:} We trained and evaluated a generalized causal model using 30,000 data samples to test its performance across different movement types.
\end{itemize}

This evaluation strategy aims to assess the practical performance and adaptability of causal graph models in complex behavior recognition tasks. It not only advances causal analysis methods for behavior and emotion computation but also provides interpretable foundations for understanding human movement.

\subsection{Evaluation and Result}
As shown in Table \ref{tab:metrics_comparison}, the Causal Graph Convolutional Network (Causal GCN) demonstrates significant effectiveness compared to baseline models across various movement types. The performance metrics—accuracy, macro precision, F1 score, and recall—provide a comprehensive assessment of the model's capabilities during both the training and testing phases, with activities such as single-leg standing, bending, and reaching forward.
In the single-leg standing task, the Causal GCN achieved high performance during the testing phase, with an F1 score of 0.883 and a recall of 0.9687, the model not only exhibits high precision in classifying the correct instances but also demonstrates a strong ability to retrieve all relevant instances. These results significantly surpass those of the baseline model, which reported an F1 score of 0.8578 and recall of 0.912. This improvement indicates that the Causal GCN effectively captures the behavioral causal relationships triggered by pain-induced protective actions. By accurately modeling the subtle joint dynamics involved in protective behaviors, the Causal GCN can better distinguish between normal and pain-affected movements.

In the bending task, the Causal GCN improved the recognition of positive samples. The recall in the training phase increased from 0.7731 in the baseline model to 0.9328 in the Causal GCN. This improvement indicates a greater ability to identify actual instances of the target behavior, thus reducing the number of false negatives. Similarly, in the complex "other" movement category, which involves a large and diverse dataset, the Causal GCN outperformed the baseline model. The training phase F1 score increased from 0.4738 to 0.5758, and recall rose from 0.3214 to 0.447. These results suggest that the model is more capable of handling complex and variable real-world data, making it more reliable for practical applications.

Across all categories, the Causal GCN consistently consistently outperformed the baseline in accuracy during both training and testing phases. This stability and high accuracy underscore its potential as a reliable model for motion behavior classification and pain detection tasks. The observed improvements in F1 score and recall in the "other" category indicate its superior ability to interpret complex motion and causal relationships. Unlike traditional models that may struggle with understanding the intricate connections between joints and behaviors, the Causal GCN excels in analyzing causal flows and using this information to make accurate predictions.

To further validate our findings, Figure \ref{fig:metrics_trend_comparison} visually illustrates the performance trends. The Causal GCN’s metrics consistently rank higher and show greater stability across all movement types compared to the baseline model. This graphical representation provides strong evidence of its overall performance advantage, reinforcing the conclusion that integrating causal structures into the model enhances both interpretability and performance.

When compared to other studies in human motion analysis and pain detection, our Causal GCN model stands out. While some previous models focus solely on improving accuracy without considering the underlying causal mechanisms, our approach not only achieves high accuracy but also offers valuable insights into the causal relationships among joints. This ability to model causal connections leads to a deeper understanding of how pain influences human motor behavior, providing a solid foundation for developing personalized pain management and treatment strategies.



\begin{figure*}[htbp]
\centering
\includegraphics[width=0.87\textwidth]{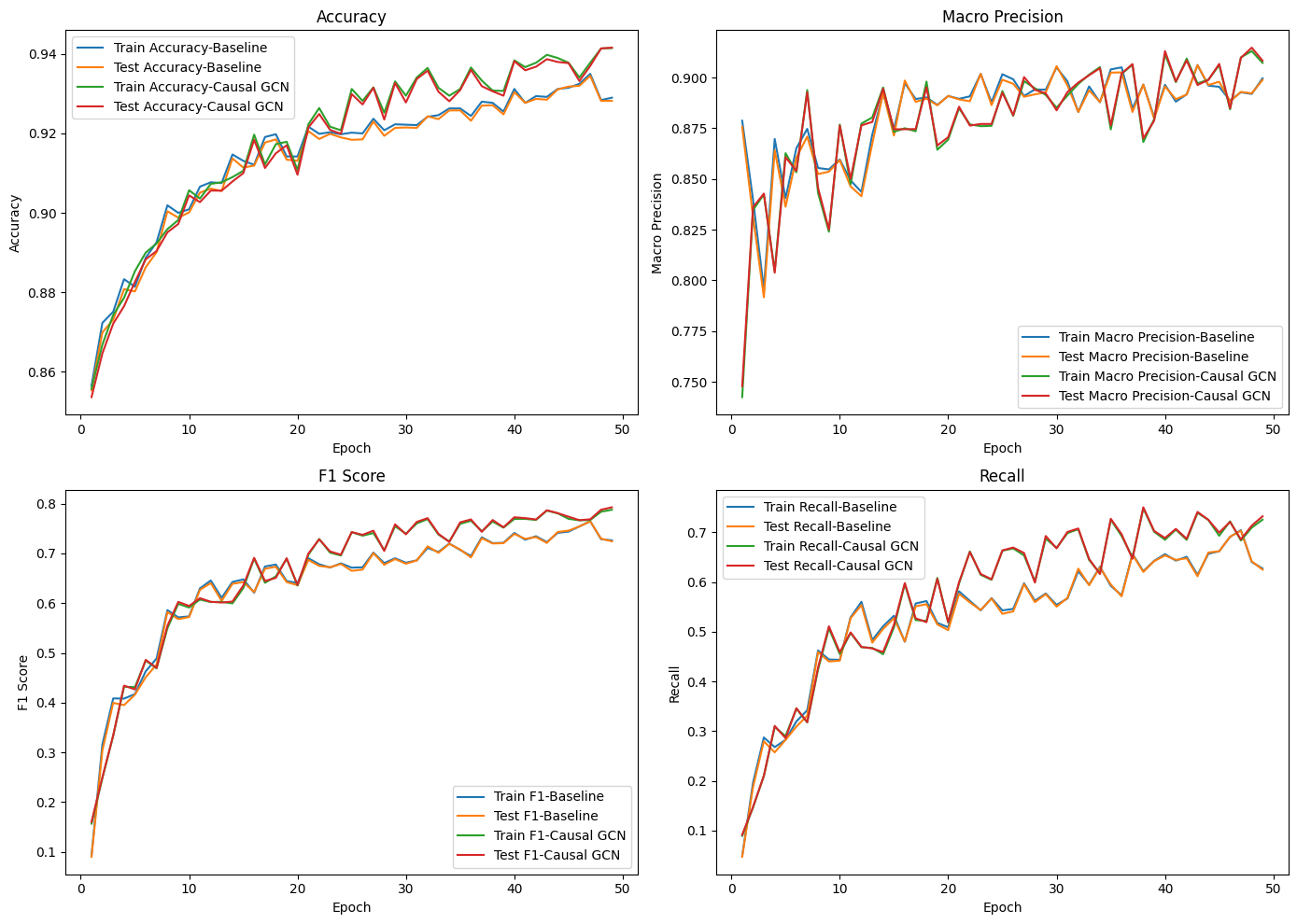}
\caption{Comparative Evaluation of Baseline and Causal GCN Performance Metrics Without Exercise Type Labeling}
\label{fig:general}
\end{figure*}

\afterpage{%
\begin{table*}[htbp]
\centering
\caption{Performance Metrics Comparison of General Models without Exercise Type Labels}
\label{tab:general_model_metrics}
\begin{tabular}{@{}lcccc@{}}
\toprule
\textbf{Model} & \textbf{Accuracy} & \textbf{Precision} & \textbf{F1} & \textbf{Recall} \\ \midrule
\textbf{Baseline - Train} & 0.929 & 0.900 & 0.726 & 0.628 \\
\textbf{Baseline - Test}  & 0.928 & 0.900 & 0.724 & 0.625 \\
\textbf{Causal GCN - Train} & \textbf{0.942} & \textbf{0.907} & \textbf{0.788} & \textbf{0.726} \\
\textbf{Causal GCN - Test}  &  \textbf{0.942} &  \textbf{0.908} &  \textbf{0.793} &  \textbf{0.733} \\ \bottomrule
\end{tabular}
\end{table*}
}


To evaluate the generalization capability of causal graph models in behavior recognition, we conducted an additional experiment comparing the performance of the baseline model with that of the general causal graph model, which did not rely on exercise type labels. As shown in Table \ref{tab:general_model_metrics}, after 50 epochs of evaluation, the Causal GCN outperformed the baseline model across all metrics. Specifically, the testing F1 score improved from 0.724 to 0.793, and the testing recall increased from 0.625 to 0.733. Figure \ref{fig:general} further illustrates the Causal GCN’s superior and more stable performance across all indicators. These results clearly demonstrate that incorporating causal structures enhances the model's ability to generalize across different behavior types, making it more adaptable to a wide range of real-world scenarios.


\section{Conclusion}
In this study, we propose an innovative causal representation learning method aimed at deepening our understanding of human movement and behavioural patterns. Utilising the EmoPain dataset, we have developed a two-stage framework combining the Peter-Clark  algorithm and KL divergence to learn representations of complex causal relationships between human joints\cite{clark1989cn2}\cite{kullback1951kullback}.

This model integrates biomechanical representation extraction into human motion analysis, enhancing machine learning models' interpretability and performance. By capturing inter-joint causal relationships, our method accurately identifies and responds to pain-induced motion behavior changes. The causal GCN model outperforms the baseline undirected GCN across multiple metrics, demonstrating its efficacy and applicability. Our approach generates highly interpretable and robust representations, surpassing traditional graph convolutional neural networks in identifying pain-induced behavioral changes. It excels in handling emotion-influenced protective behaviors and shows strong generalization across different movement types, enhancing practical robustness. The innovation lies in capturing subtle causal relationships overlooked by traditional methods, providing deeper insights into human behavior analysis. This causal representation learning method offers a novel approach to understanding complex human motion patterns and their underlying mechanisms.

Our research not only advances the integration of machine learning and personalised medicine but also opens new avenues for personalised adaptive healthcare and behavioural analysis. By introducing causal inference into representation learning, our method lays the foundation for developing more precise and adaptive intelligent healthcare solutions. This approach holds immense potential in improving the interpretability and personalised adaptability of human-centred computational models.

Looking ahead, this research prides new ideas for constructing more intelligent and humanised medical systems. It not only helps improve the accuracy of personalised behavioural analysis but may also find broad applications in related fields such as affective computing and rehabilitation medicine. We believe that this representation learning method, integrating causal inference, will open new prospects for the application of artificial intelligence in healthcare, driving further development of personalised medicine.

\section{Discussion}

This study introduces a causal representation learning framework to advance human motion and behavior analysis. Integrating causal inference with representation learning, we propose an interpretable two-layer model: at the macroscopic level, the Peter-Clark algorithm constructs an initial causal graph of joint relationships; at the microscopic level, KL divergence quantifies information flow between joints \cite{gu2024advancing}. This structure not only extracts features from complex human data but also enables causal traceability, enhancing clinical interpretability \cite{ahmad2018interpretable, jin2022explainable}.

By uncovering deeper temporal and spatial dependencies beyond statistical correlations, the method improves model accuracy and provides biomechanical explanations \cite{lee2010kinematic}. Future work will focus on time-varying causal discovery and multimodal fusion to expand medical diagnostic applications. Building on this framework, we further analyzed causal differences between protective and non-protective behaviors. Results on the EmoPain dataset aligned with medical theories: chronic pain patients exhibited stronger hip-knee causal links during bending tasks (Vlaeyen’s model) \cite{vlaeyen2000fear}, and heightened trunk-limb coupling during sit-to-stand in high pain catastrophizers \cite{sullivan2001theoretical}. The framework also validated Hodges and Tucker’s motor adaptation model through complex spinal joint networks \cite{hodges2011moving}, and quantified emotion-pain interactions, showing depression-related reductions in joint causality \cite{asmundson2012role, de2014mindfulness}.

While effective, the framework assumes clean, segmented data and relies on KL divergence, which remains sensitive to noise and does not confirm true causality. Discrete-time causal graphs further limit modeling of dynamic dependencies. Future directions include adapting causal learning to high-frequency, multimodal settings such as affective computing and human-robot interaction, with a focus on scaling, interpretability, and mitigating bias—key steps toward trustworthy, human-centered AI systems.

\section*{Acknowledgment}

I would like to express my sincere gratitude to Prof. Nadia Berthouze for her invaluable guidance and thoughtful suggestions throughout this work. I am also deeply appreciative of the support and guidance provided by the University College London Interaction Centre, University College London, UK and The Future Laboratory, Tsinghua University, Beijing, China.

\clearpage
\section*{ETHICAL IMPACT STATEMENT}
Our research leverages causal inference in affective computing to enhance pain recognition, with a focus on understanding human joint dynamics and complex behaviors. This ethical impact statement highlights key considerations and mitigation strategies in line with FG 2025 guidelines.

\subsection{Potential Risks and Harms}
\begin{itemize}
    \item \textbf{Individual Harm:} Misclassification of pain behaviors, particularly protective movements, could lead to inadequate medical responses or misinterpretation of pain levels, posing risks to patient safety. Additionally, handling sensitive health data raises concerns around privacy, as potential breaches could expose individuals to discrimination or unwanted disclosure of medical conditions.
    \item \textbf{Societal Impact:} There is a risk that this technology, if not equitably designed, could perpetuate disparities in healthcare, especially if the model's performance varies across demographic groups. Furthermore, reliance on automated assessments of complex behaviors may diminish the role of human expertise, potentially impacting the personalization of healthcare.
\end{itemize}

\subsection{Risk-Mitigation Strategies}
Our study incorporates the following strategies to mitigate these risks:
\begin{itemize}
    \item \textbf{Equitable Representation and Fairness:} We utilize the EmoPain dataset \cite{aung2015automatic}, ensuring diverse participant representation to enhance model performance across demographic groups and reduce biases in pain recognition, particularly in detecting protective behaviors.
    \item \textbf{Privacy and Security:} Robust privacy protocols, including data anonymization and secure storage, are implemented to protect sensitive health information and prevent unauthorized access.
    \item \textbf{Explainability and Model Interpretability:} By employing causal inference with the Peter-Clark (PC) algorithm and Kullback-Leibler (KL) divergence, our framework generates interpretable representations of joint dynamics, providing clinicians with insights into how the model interprets complex behaviors, which supports informed clinical decision-making.
    \item \textbf{Human-Centered Integration:} Our approach emphasizes a supportive role for this technology, designed to augment clinical judgment rather than replace it, ensuring that healthcare providers retain control and can effectively interpret model outcomes.
    \item \textbf{Validation Across Demographics:} Continuous performance validation across demographic groups ensures that the model remains robust and reliable, reducing risks of differential accuracy in pain recognition.
\end{itemize}

\subsection{Benefit-Risk Analysis}
The potential benefits include more accurate and objective assessments of pain-related behaviors, contributing to improved patient outcomes and advancements in personalized pain management. We believe these benefits outweigh the risks due to the following reasons:
\begin{enumerate}
    \item \textbf{Mitigation Measures:} Our comprehensive risk-mitigation strategies address primary concerns around accuracy, privacy, and fairness.
    \item \textbf{Supplementary Role in Healthcare:} The model is intended to assist, not replace, healthcare professionals, ensuring that clinical expertise remains integral to patient care.
    \item \textbf{Impact on Healthcare Innovation:} Improved detection of protective behaviors and pain responses enhances the capacity for tailored healthcare solutions, offering societal benefits and contributing to healthcare advancements.
\end{enumerate}

\subsection{Human Subjects Protection}
All data utilized in this study is derived from the EmoPain dataset \cite{aung2015automatic}, collected under ethical oversight with participants' informed consent. The data remains anonymized and is managed under stringent security protocols, safeguarding participant privacy and maintaining data integrity throughout the research process.
\newpage 
\bibliographystyle{IEEEtran}
\bibliography{reference}

\vspace{12pt}
\clearpage
\appendix
\section{Detailed Proof of Theorem}
\subsection{Proof of Theorem 1}
\label{proof1}
\begin{proof}
Let $(i, j) \in E^*$ be an edge in the CPDAG $\mathcal{G}^*$ output by the PC algorithm.

If there is a causal relationship from $i$ to $j$:
\begin{itemize}
    \item[a)] The PC algorithm will not remove this edge during the conditional independence tests.
    \item[b)] The KL divergence $D_{KL}(i \rightarrow j)$ will be positive:
    \[
    D_{KL}(i \rightarrow j) = \sum_{x_j \in \mathcal{X}_j} P(x_j \mid x_i) \log\left(\frac{P(x_j \mid x_i)}{P(x_j)}\right) > 0
    \]
    \item[c)] Therefore, in the representation $\phi(v_i)$:
    \[
    \phi(v_i)_j = \frac{D_{KL}(i \rightarrow j)}{\sum_{k \in \text{Ne}(i)} D_{KL}(i \rightarrow k)} > 0
    \]
\end{itemize}

Conversely, if there is no causal relationship from $i$ to $j$:
\begin{itemize}
    \item[a)] The PC algorithm will remove this edge during the conditional independence tests.
    \item[b)] Since $j \notin \text{Ne}(i)$, $\phi(v_i)_j$ will be undefined (effectively zero).
\end{itemize}

Thus:
\[
\phi(v_i)_j > 0 \iff (i, j) \in \mathcal{G}^*
\]
\end{proof}

\subsection{Proof of Theorem 2}
\label{proof2}
\begin{proof}
Let $\alpha > 0$ be a scaling factor applied to the data.

For any two probability distributions $P$ and $Q$, KL divergence is invariant under scale transformations:
\[
D_{KL}(\alpha P || \alpha Q) = \sum_{x} \alpha P(x) \log\left(\frac{\alpha P(x)}{\alpha Q(x)}\right)
\]
\[
= \alpha \sum_{x} P(x) \log\left(\frac{P(x)}{Q(x)}\right)
\]
\[
= \alpha D_{KL}(P || Q)
\]

In our representation $\phi$:
\[
\phi(v_i)_j = \frac{D_{KL}(i \rightarrow j)}{\sum_{k \in \text{Ne}(i)} D_{KL}(i \rightarrow k)}
\]

Under a scale transformation by $\alpha$:
\[
\phi'(v_i)_j = \frac{\alpha D_{KL}(i \rightarrow j)}{\sum_{k \in \text{Ne}(i)} \alpha D_{KL}(i \rightarrow k)}
\]
\[
= \frac{\alpha D_{KL}(i \rightarrow j)}{\alpha \sum_{k \in \text{Ne}(i)} D_{KL}(i \rightarrow k)}
\]
\[
= \frac{D_{KL}(i \rightarrow j)}{\sum_{k \in \text{Ne}(i)} D_{KL}(i \rightarrow k)}
\]
\[
= \phi(v_i)_j
\]

Therefore, $\phi'(v_i)_j = \phi(v_i)_j$ for all $i, j$, proving that $\phi$ is invariant to scale transformations of the data.

\end{proof}

\section{EmoPain Dataset DAGs}
\begin{figure*}[htbp]
  \centering
  \includegraphics[width=\textwidth]{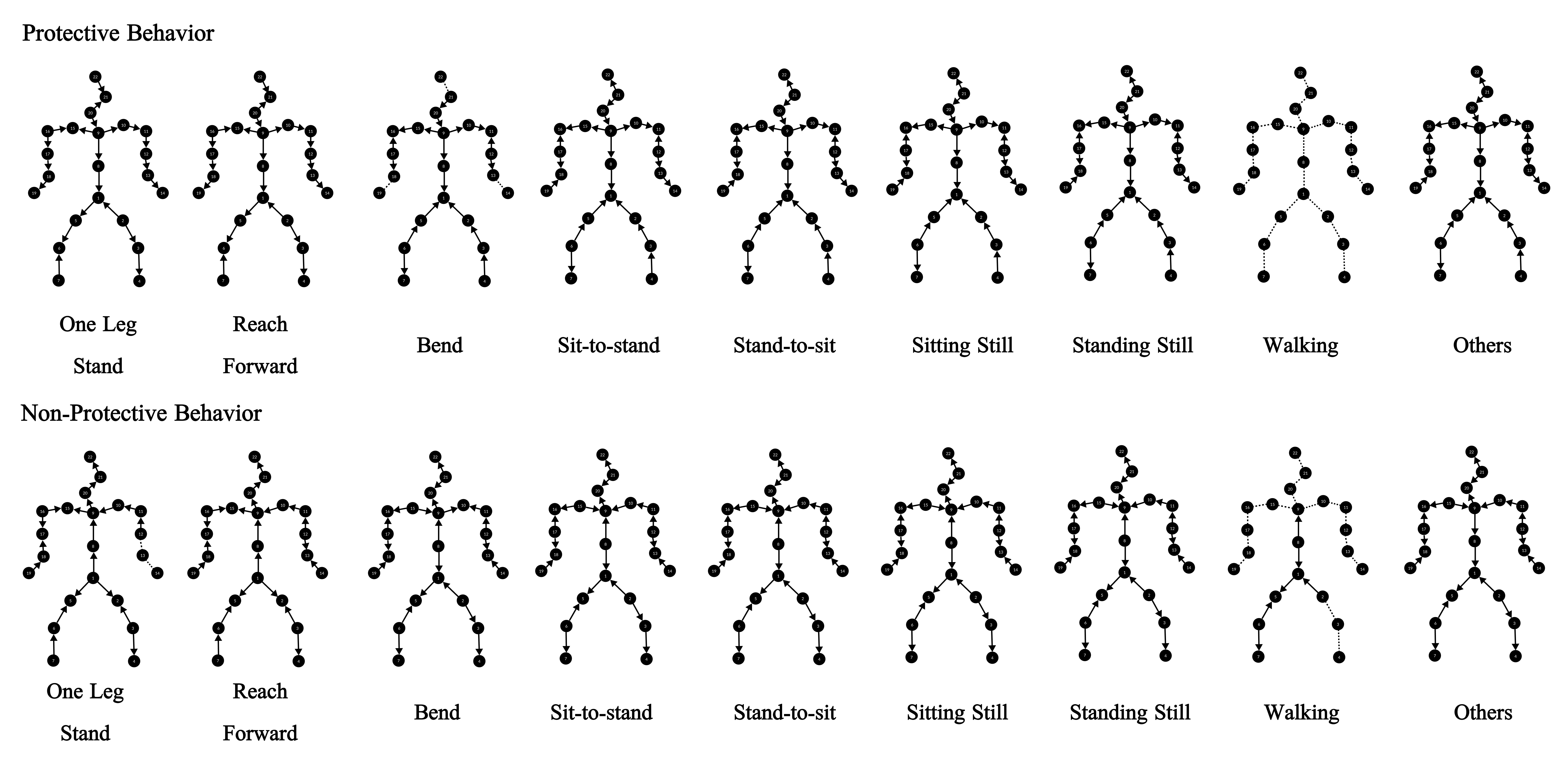}
  \caption{Causal graph networks for all movement types in the EmoPain dataset, categorized into protective and non-protective behaviors. Each graph highlights the joint-level causal relationships for movements such as One Leg Stand, Reach Forward, and others. Protective behaviors demonstrate stronger stabilization patterns, while non-protective behaviors exhibit more freedom of movement.}
  \label{fig:emopain_dags}
\end{figure*}

\newpage
\begin{table*}[htbp]
\centering
\caption{CLBP Participants' Profile Summary}
\label{tab:clbp_participants}
\begin{tabular}{cccccccc}
\hline
Age & Gender & HADS Score & PCS Score & Self Report Pain N & Self Report Pain D & Self Report Anxiety N & Self Report Anxiety D \\ \hline
63 & M & 4 & 2 & 0 & 0 & 0 & 0 \\
53 & F & 25 & 14 & 0 & 0.2 & 0 & 0 \\
65 & F & 16 & 13 & 5.5 & 5.8 & 0.9 & 0.9 \\
27 & F & 25 & 18 & 5.1 & 5.7 & 1.9 & 3.5 \\
31 & F & 8 & 2 & 2.8 & 2.7 & 0 & 0 \\
64 & M & 20 & 17 & 5 & 5.6 & 1.9 & 1.7 \\
62 & M & 25 & 30 & 5.8 & 6.7 & 0 & 0 \\
56 & M & 11 & 12 & 3.9 & 4.7 & 0 & 0 \\
36 & M & 19 & 15 & 1.4 & 1.8 & 0 & 0 \\
58 & F & 17 & 13 & 0.4 & 0.8 & 0 & 0 \\
- & F & 8 & 6 & 6.1 & 3.9 & 0 & 0 \\
55 & F & 11 & 15 & 1.1 & 1.7 & 1.3 & 2.1 \\
33 & F & 11 & 8 & 4.1 & 3.9 & 2.9 & 2.3 \\
19 & M & 30 & 42 & 7.1 & 7.6 & 2.9 & 2.7 \\
38 & F & 5 & 0 & 0 & 0 & 0 & 0 \\
- & F & 21 & 37 & 2.6 & 3 & 0 & 0 \\
51 & F & 15 & 5 & 0.5 & 0.1 & 0.1 & 0 \\
67 & M & 24 & 33 & 6.6 & 8.7 & 6.3 & 8 \\
62 & F & 8 & 11 & 1.1 & 1.4 & 0.1 & 0.3 \\
56 & F & 32 & 44 & 4.7 & 5.6 & 4 & 2.3 \\
65 & F & 11 & 17 & 0 & 0 & 0 & 0 \\
50 & F & 34 & 42 & 6.1 & 7.7 & 0 & 0 \\ \hline
50.5 & 17.3 & 18 & 3.18 & 3.53 & 1.01 & 1.08 \\ \hline
\end{tabular}

\vspace{0.2cm} 
{\footnotesize \textit{Note:} The scores shown are: sum of Hospital Anxiety and Depression Scores (HADS, scale: 0-42), sum of the Pain Catastrophizing Scores (PCS, scale: 0-52) and mean levels of Self Reported Pain and Anxiety for all exercises in the normal (N) and difficult trials (D) (scale: 0-10). The final row contains the mean age and scores. These demographic information of patients in the EmoPain dataset is taken from page 6 of the reference article for the EmoPain dataset: "The Automatic Detection of Chronic Pain-Related Expression: Requirements, Challenges and the Multimodal EmoPain Dataset. \cite{aung2015automatic}}
\end{table*}

\section{GNN Model}
\begin{figure*}[htbp]
  \includegraphics[width=1\textwidth]{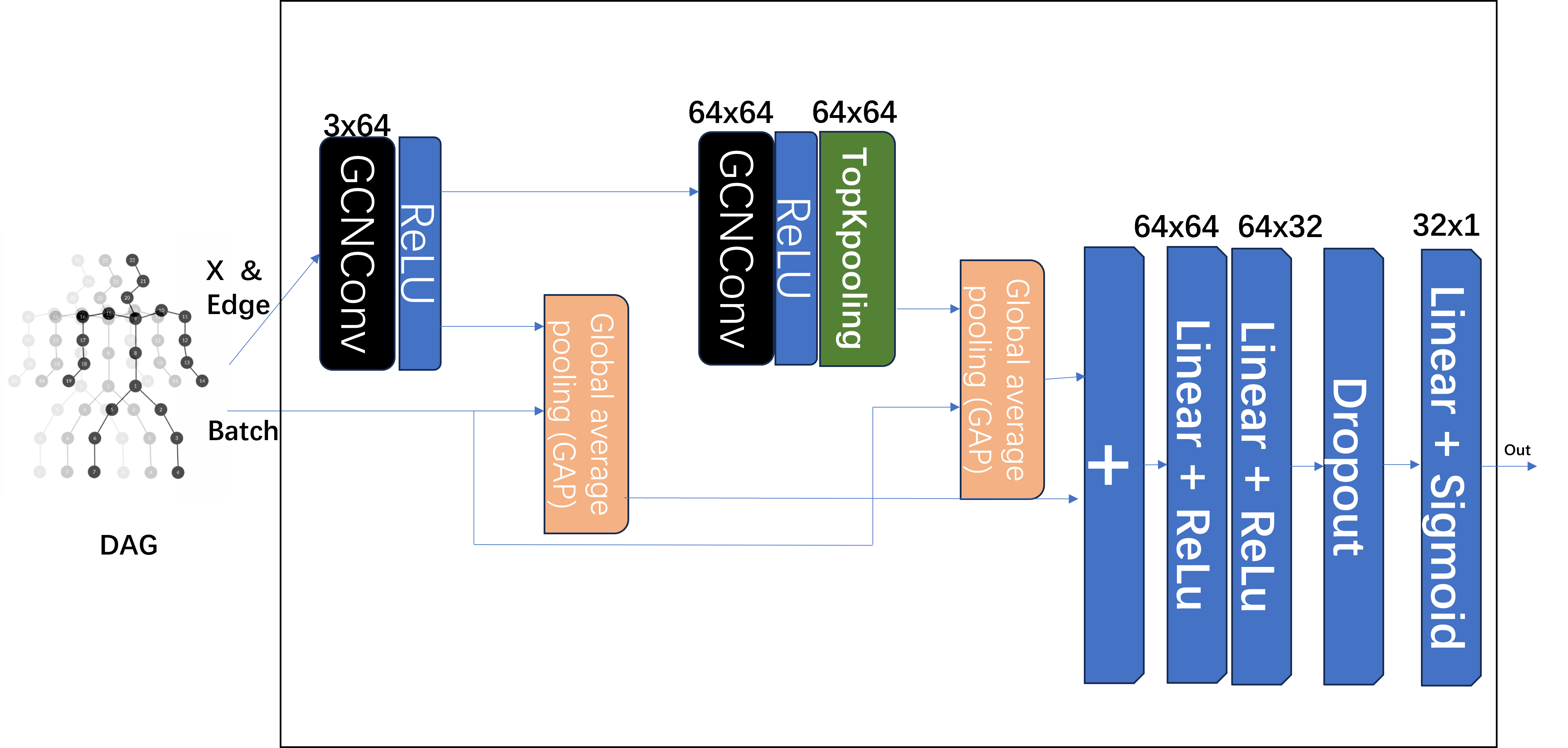}
  \caption{Architecture of the Graph Convolutional Network (GCN) Model}
  \label{fig:gcnarticture}
\end{figure*}
\end{document}